\documentclass[10pt, conference, letterpaper]{IEEEtran}

\usepackage{cite}

\usepackage{graphicx}
\ifCLASSINFOpdf
\else
\fi
\begin{document}

\title{Deep Learning Radio Frequency Signal\\Classification with Hybrid Images}

\author{\IEEEauthorblockN{Hilal Elyousseph}
\IEEEauthorblockA{Department of Electrical Engineering\\
King Saud University\\
Riyadh, Saudi Arabia\\
439106893@student.ksu.edu.sa}
\and
\IEEEauthorblockN{Majid L Altamimi}
\IEEEauthorblockA{Department of Electrical Engineering\\
King Saud University\\
Riyadh, Saudi Arabia\\
mtamimi@ksu.edu.sa}
\and
}

\maketitle

\begin{abstract}
In recent years, Deep Learning (DL) has been successfully applied to detect and classify Radio Frequency (RF) Signals. A DL approach is especially useful since it identifies the presence of a signal without needing full protocol information, and can also detect and/or classify non-communication waveforms, such as radar signals. In this work, we focus on the different pre-processing steps that can be used on the input training data, and test the results on a fixed DL architecture. While previous works have mostly focused exclusively on either time-domain or frequency domain approaches, we propose a hybrid image that takes advantage of both time and frequency domain information, and tackles the classification as a Computer Vision problem. Our initial results point out limitations to classical pre-processing approaches while also showing that it's possible to build a classifier that can leverage the strengths of multiple signal representations.
\end{abstract}


\IEEEpeerreviewmaketitle
\section{Introduction}
Being able to classify unknown RF signals is an important topic, such as for cognitive radio and defense applications, and should remain open to techniques of the future, such as DL. As DL has been successfully applied to image classification, object detection, and speech prediction, etc., it has also been beneficial in detecting and classifying RF transmissions. RF transmissions can be thought of in a cooperative context, such as being the intended recipient of a transmission, but also in a non-cooperative context, such as detecting the presence of an unauthorized transmission. In a non-cooperative context, prior information about the signal type cannot always be assumed, which means looking for specific features, preamble, pilot tones, etc. cannot be used to help detect and classify the signal. Such a scenario is also called blind detection, which is especially important in a military application where the goal is to achieve spectral awareness\cite{9044593}.

Using DL to identify the presence of RF signals is a fairly new approach, and it can be applied to both commercial and military domains. In the context of cognitive radio, for example, DL has been used in the form of a Convolutional Neural Network (CNN) to identify the presence of a Radar waveform against a background of Wi-Fi and LTE signals\cite{8254105}. This approach allows a low-cost device to make a decision about the specific presence of a signal from the physical layer, which reduces control overhead\cite{wu2020automatic}. The end result is that the listening device, termed a secondary user, will share the RF spectrum with the radar waveform and only transmit when it identifies the waveform is not present. From a security perspective, DL has also been used in the form of a Dense Neural Network (DNN) to both detect the presence of a drone signal and to also distinguish between drones from different manufacturers, which can be important if the counter-measure used depends on the type of drone detected\cite{allahham2019dronerf}. DL has also been used to classify unknown modulation types from the raw time-domain samples of an RF signal, which is important both to build software defined radio (SDR) networks that can quickly adapt to different modulation schemes, or as a first step in a system that provides more information about what type of RF transmissions are occurring in a certain area\cite{o2016convolutional}.

The main motivation of our work started from the observation that the classification of drone types and specific wireless technologies have mostly focused on frequency-domain representations of the RF signals\cite{allahham2019dronerf,behura2020wist,schmidt2017wireless,8815481,ozturk2020rfbased,grunau2018multilabel,basak2020drone,8357902}, while classifying unknown modulation schemes has focused on time-domain representations\cite{wu2020automatic,8357902,o2016convolutional,9128408,o2018over}. This is because distinguishing between many modulation schemes requires the phase information of the signal, which gets lost in common frequency domain representations such as the Power Spectral Density (PSD) and the Short-Time Fourrier Transform (STFT). To classify different drone models however, it has been shown that time domain envelopes are more easily lost in background noise, making frequency domain representations more suitable to successfully classify the drone models\cite{ozturk2020rfbased}. This shows the need to investigate a hybrid approach that uses both time domain and frequency domain representations of RF signals to classify both unknown modulation schemes and also specific technology identification. Using a fixed CNN architecture, we test the model's ability to predict various RF signals after different pre-processing steps, which shows the limitations of classical approaches and how hybrid pre-processing steps are required for more robustness.

In the next section we will provide a brief literature review with the focus on pre-processing steps used for signal classification with DL. In Section \ref{methodology}, we outline our experimental setup. In Section \ref{results}, we discuss our results, and finally conclude the work in Section \ref{conclusion}. 

\section{Literature Review}
\label{literature}
In the past, most signal classification techniques have required advanced domain knowledge and the ability to extract features of identified signals in order to classify them. This can include statistical moments\cite{soliman1992signal} and/or spectral correlation\cite{fehske2005new}, among other techniques. A survey that goes over the traditional techniques can be found in \cite{dobre2007survey}. RF Signal classification with DL on the other hand requires less expert knowledge, since CNNs and other DL techniques can be used to extract features from raw data. We will show however that there still needs to be special attention paid to the pre-processing steps. 

RF signal classification with DL has mostly been studied in the context of either Automatic Modulation Classification (AMC) \cite{o2018over, 9128408, chen2020radio} or Wireless Interference Identification (WII) \cite{behura2020wist, schmidt2017wireless, 8815481}. These are essentially the same tasks with the difference being the signals involved. Work in AMC has mostly been using time-domain information, while work in WII has been performed using frequency domain representations. The authors in \cite{8325299} performed experimental work comparing the results of a CNN used for AMC, and then again separately for WII. They showed that frequency-domain representations outperform time-domain representations for WII due to the signals involved having more distinct features in the frequency domain, although both time and frequency representations successfully performed WII. 

AMC is less forgiving in this regards, since certain modulation schemes are difficult to distinguish if we lose the phase information, which is the case with popular frequency domain representations. This is precisely why AMC works have focused on time-domain representations. A popular time-domain representation of an RF signal is to use the IQ values, which are a complex pair representing the amplitudes of the in-phase and quadrature components of the signal in rectangular coordinates. Another popular time-domain representation is to convert the IQ values to polar coordinates to have an amplitude-phase representation. The authors of \cite{8357902} attempted this representation on the dataset of \cite{o2016convolutional}, using the same CNN architecture, but found similar results to using IQ values. Significantly, they also attempted a Short Term Memory (LSTM) architecture with IQ representation but the model was not able to learn at all. The same LSTM model however with amplitude-phase signal representations achieved better results than the IQ representation used with the CNN. This shows the importance of considering both pre-processing steps and DL architecture, since certain DL architectures might fail to learn, depending on the pre-processing steps.

Another pre-processing step to consider is that these time-domain representations can be segmented and kept as time-series vectors as performed in many papers\cite{o2016convolutional, 8325299, 9128408, o2018over}, or these segments can instead be plotted and used to train an image classifier as performed in \cite{ozturk2020rfbased}. 


For WII, frequency domain representations have performed better since the frequency components of a signal remain more distinctive even under lower SNR. WII can appear as different use cases, such as classifying different UAVs\cite{allahham2019dronerf, ozturk2020rfbased, 8913640, basak2020drone}, detecting/classifying different Radar Waveforms \cite{8254105, akyon2018deep}, identifying specific technologies in the ISM band\cite{8815481}, securing Global Navigation Satellite Systems (GNSS) signals \cite{data5010018}, and more. Like time-domain representations, frequency-domain representations can also be kept as a vector of values (such as DFT coefficients) used in \cite{allahham2019dronerf} or further transforms can be applied to make a PSD image, or STFT image (also known as a spectrogram), used in \cite{ozturk2020rfbased, basak2020drone} and others. A benefit of using STFT images is that a time dependency can be captured which allows such a system to be used to detect frequency hopping signals such as in \cite{8979431}. In Fig. 1,  we summarized the classical pre-processing steps used in these different works.


\begin{figure}[!t]
\center
\includegraphics[width=\linewidth]{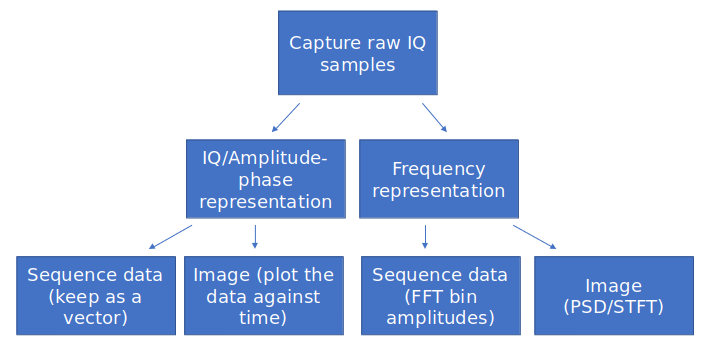}
\caption{Classical pre-processing steps before training a Deep Learning Model.}
\label{fig1}
\end{figure}

\section{Methodology}
\label{methodology}

In this section, we present our approach to construct and demonstrate the proposed DL system. That includes generating the experimental data, describing the pre-processing steps in detail, and showing proposed representation of the signal and the DL network architecture. Without loss of generality, we demonstrate our approach on BPSK, QPSK, 16-QAM, and GFSK signals to have a sufficient variety that would be either easy or difficult to discern based on the pre-processing steps. These signals would emulate building a RF signal detector that can alert for certain signals, while declaring other signals as safe/authorized transmissions. Also, in a real world setting there would be more advanced techniques, such as multi-carrier, frequency hopping, and time/frequency multiplexing, but this will of course depend on the use case when designing a DL based classifier. 

 \subsection{Data Generation}
 
\begin{figure}[!]
\center
\includegraphics[width=\linewidth]{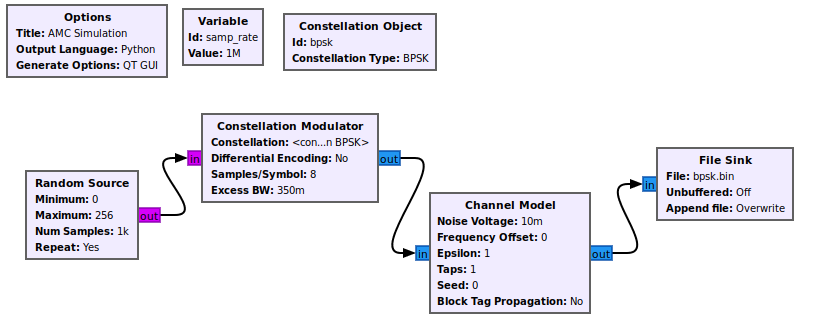}
\caption{GNU Radio flowgraph to generate the signal recordings and add channel effects.}
\label{fig2}
\end{figure}


In the literature, we found the datasets of captured signals in the real world for the purpose of academic usages is limited. Moreover, the variety of the signals in some publicly available datasets is narrow. Therefore, we used the GNU Radio platform to simulate real world signals captured for different modulation schemes with noise. The actual modulation schemes were chosen so that the classification accuracy would depend on the pre-processing steps used. 

Shown in Fig. 2, each signal was simulated with random data at a sampling rate of one million samples per second. The result was four different signal recordings, that where then clipped to all be the same length. We then segmented the signals into complex 1D vectors of length [1x512], and kept 50 segments for each signal type. For four signal types this gave us 200 total RF segments for classification purposes.

\subsection{Classical Pre-Processing Steps}

\begin{figure}[!]
\centerline{\includegraphics[width=\linewidth, height=5.5cm]{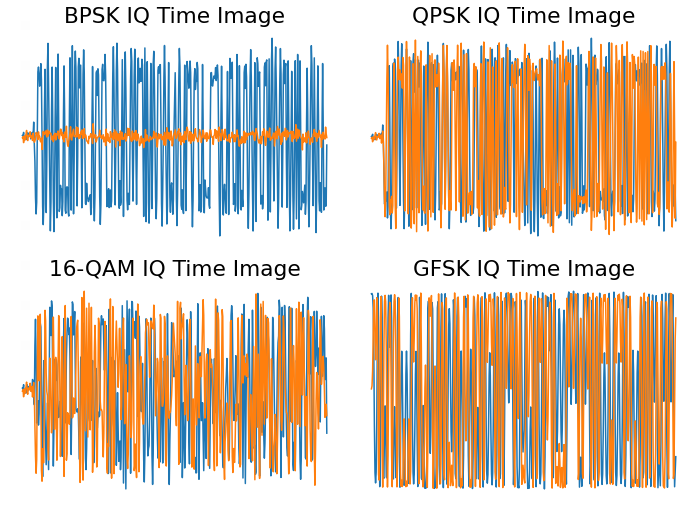}}
\caption{Raw IQ time-series images. The in-phase component is shown in blue and the quadrature component is in orange.}
\label{fig3}
\end{figure}

We followed the flowgraph of Fig. 1, namely turning the recorded signals into their time and also frequency representations. The time domain images show the in-phase and quadrature components of our signal. For the time representation tests we normalized the data between negative one and one, and then plotted the amplitudes of the in-phase and quadrature components as an image, shown in Fig. 3. Looking at the images, we see the BPSK signal has steep amplitude transitions from one to negative one, which is how the scheme conveys binary data. QPSK uses the same technique but with both the I and Q components. The 16-QAM time image has a staggered appearance instead of a steep vertical transitions, and this is due to 16-QAM having more possible states which allows it to transmit more bits per symbol. Finally, GFSK modulates a carrier wave with binary data by changing the frequency of the carrier wave, which will show distinct peaks in the frequency domain. The methods each modulation scheme uses to vary its amplitude and/or phase is how the time domain images allows us to distinguish between the signals, but as there is increasing noise or impairments it will become difficult to visually distinguish between them.

\begin{figure}[!]
\center 
\includegraphics[width=\linewidth]{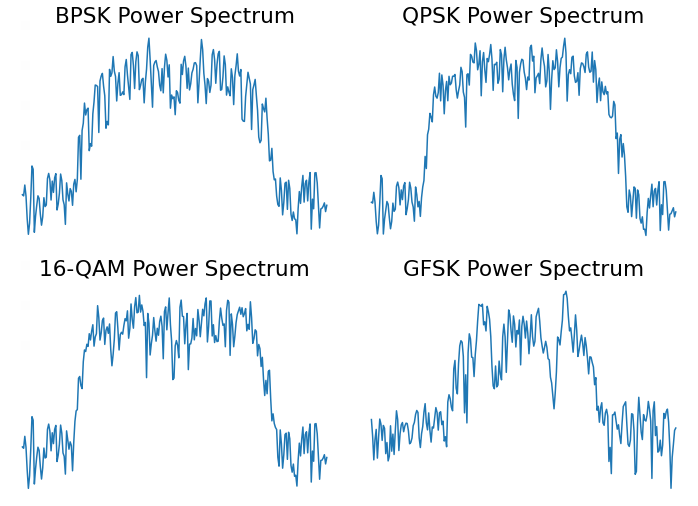}
\caption{PSD images. The vertical axis represents power density in dB and the horizontal axis is the discrete frequency bin. These are the actual images used to train the model.}
\label{fig4}
\end{figure}

For the frequency domain representations, we used both PSD images shown in Fig. 4, and STFT images shown in Fig. 5, to train our models. Both PSD and STFT transforms were implemented with the Python library Matplotlib. The PSD images estimate how much power is in each frequency bin, where we chose 256 frequency bins. The STFT is derived by first segmenting the total length of samples into smaller segments, and calculating the PSD of each segment before combining the results. This allows us to see how the frequency content is changing over time. From our choice of signals, it's clear that the differences between the BPSK, QPSK, and 16-QAM signals are not as obvious in the frequency domain.

\begin{figure}[!]
\center 
\includegraphics[width=\linewidth]{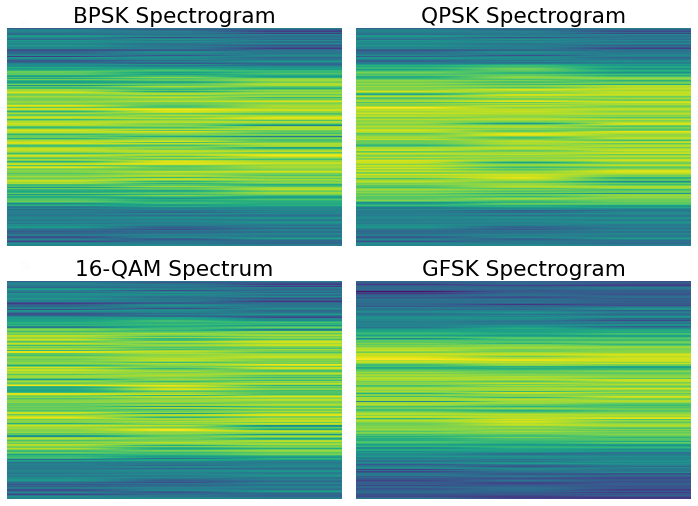}
\caption{Spectrogram Images with frequency in the vertical axis and time in the horizontal axis.}
\label{fig5}
\end{figure}

\begin{figure}[htbp]
\center
\includegraphics[width=\linewidth]{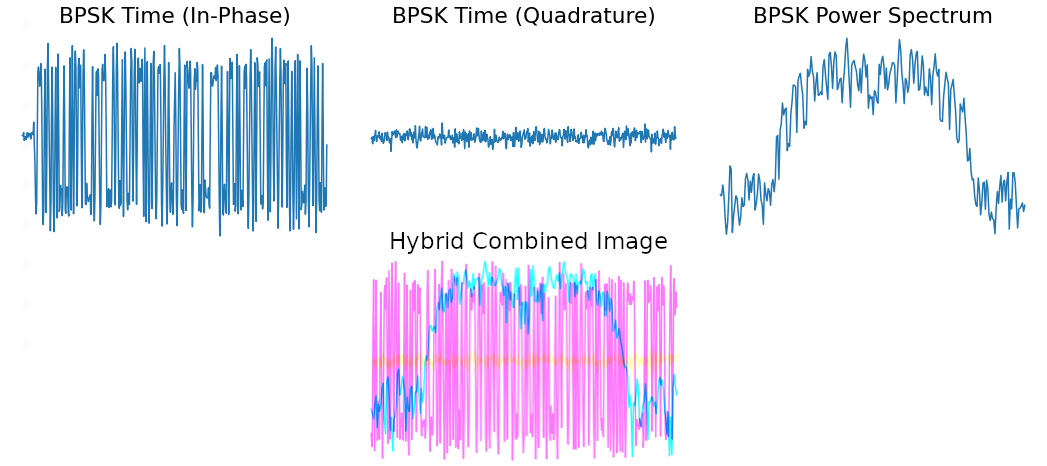}
\caption{Combining the in-phase, quadrature, and PSD images as separate channels to form one RGB image, shown in the bottom.}
\label{fig6}
\end{figure}

\subsection{Proposed representation}
It has been shown in other RF classification works that gray scale images will have the same results as RGB images\cite{8254105}. This opens up an opportunity for us since we only need one channel to represent a given input. We used the in-phase component time image as the R channel, the quadrature component time image as the G channel, and the PSD image as the B channel, to form a hybrid three channel image, shown in Fig. 6. The image will not make much sense to the human eye, but the reason this will work for the CNN is because the filters will have a depth of three (the same as our RGB input) and each layer will extract features from each channel separately, before combining the weights to a single activation/feature map. Training a single model that can leverage the strengths of the time-domain representations used in AMC works and the noise-robustness of the frequency representations of the WII works will make for a detector/classifier that can handle different scenarios instead of treating AMC and WII as separate problems. 

\subsection{Neural Architecture}
We implemented our CNN network using Tensorflow/Keras, with the architecture shown in Table I; where all filters has a size of 3 by 3. We fixed the same architecture to test all different pre-processing steps for the purpose of a fair comparison. For all models, we used a batch size of 32 and trained for 20 epochs. We used the Adam optimizer and a categorical cross-entropy loss function. For our proposed hybrid images, we will show in the results section that our CNN learned the unique features in each channel of the image, which allowed it to obtain results that none of the classical pre-processing steps were able to achieve.


\begin{table}[!h]
\caption{CNN Architecture}
\begin{center}
\begin{tabular}{|c|c|c|}
\hline
Layer & Size & Activation \\
\hline \hline
 Conv2D & 16 Filters & ReLU \\ 
 Max Pooling 2D & 2x2 & -- \\ 
 Conv2D & 32 Filters & ReLU \\ 
 Max Pooling 2D & 2x2 & -- \\ 
 Conv2D & 64 Filters & ReLU \\ 
 Max Pooling 2D & 2x2 & -- \\ 
 Dense & 256 Neurons & ReLU \\
 Dense & 4 Neurons & Softmax \\
\hline
\end{tabular}
\label{tab1}
\end{center}
\end{table}

\begin{figure}[!]
\includegraphics[width=\linewidth]{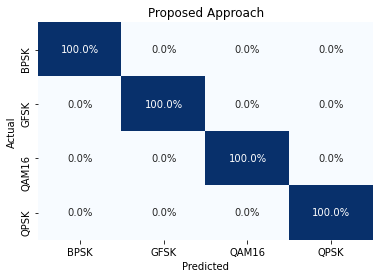}
\caption{Confusion Matrix for proposed representation.}
\label{fig8}
\end{figure}

\begin{figure*}[!]
{\includegraphics[width=\textwidth]{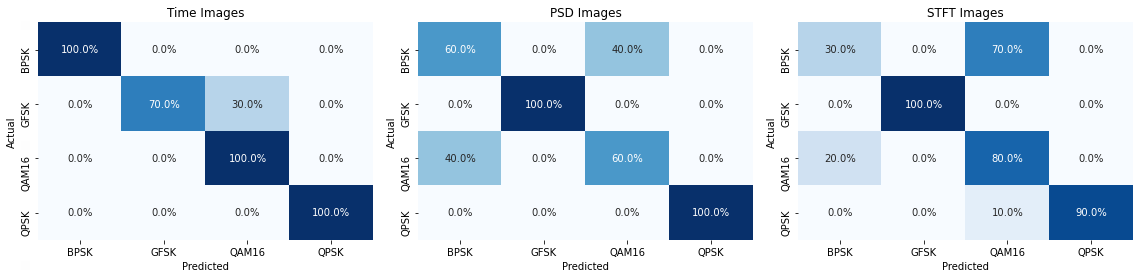}}
\caption{Confusion Matrices for the first experiments, testing the results of the Classical pre-processing steps.}
\label{fig7}
\end{figure*}


\section{Results}
\label{results}
Using the pre-processing steps discussed prior, we used GNU Radio to add noise to our signals, trained four different models with a fixed CNN architecture, an 80/20 training validation split, and tested against a separate directory of images. The final scores for the different pre-processing steps as are follows: 92.5\% accuracy for the time domain images, 80\% accuracy for the PSD images, and 75\% accuracy for the spectrograms, and finally 100\% for the proposed approach. The Confusion Matrix for our proposed approach is in Fig. 7.

For our time domain tests, the model performed well because it's easy to see how the most of the images are different from each other as in Fig. 3. However, the addition of noise made the GFSK time domain image difficult to distinguish from that of the 16-QAM signal.
For our PSD and spectrogram tests, what is most relevant to our discussion is that using frequency domain representations had poorer overall results, even though we are using the same signal recordings. The exception of course was no confusion between the GFSK signal with the other signals, and this point is precisely what allows our hybrid images to achieve full accuracy. It was clear from the frequency representations in Figs. 4 and 5 that our model trained on frequency images would have difficulty between the BPSK, QPSK, and 16-QAM signals, but not the GFSK signal since it has two distinct peaks that are obvious only in the frequency domain.

Using our proposed representation, we were able to achieve full accuracy for our test set, as shown in Fig. 7. This shows that the CNN used the information in the R and G channels (In-phase and Quadrature time images) to help distinguish between BPSK, QPSK, and 16-QAM, and that it also used the B channel (PSD images) to allow it to distinguish GFSK from the other signals. The proof for this is that neither the time domain images nor frequency domain images achieved 100\% accuracy on their own, as shown in Fig. 8. 

\section{Conclusion}
\label{conclusion}
In this work, we tested a fixed CNN architecture with several proposed pre-processing steps using the same signal datasets. We showed that a CNN fed with multiple representations of a signal will learn what it needs to minimize its loss function, and thus take advantage of the strengths of the different channels in our proposed hybrid image. Where most other works either focused on AMC or WII as separate problems, we showed that due to CNN filters extracting features from each image channel individually, we can use hybrid images to allow us to classify RF signals even if they only have more distinctive features in the frequency domain as opposed to the time domain, or vice versa. 

The promising preliminary results opens research direction to examine the proposed method for more signals types and for different SNR values. Moreover, testing the proposed method against time-frequency transforms that maintain phase information, is also another potential research area.

\bibliographystyle{IEEETran}
\bibliography{main}

\end{document}